\documentclass[11pt,a4paper]{article}
\usepackage[hyperref]{acl2020}
\usepackage{times}
\usepackage{latexsym}
\usepackage{graphicx}
\usepackage{enumitem}

\usepackage{microtype}

\aclfinalcopy 

\title{MALM: Mixing Augmented Language Modeling for Zero-Shot Machine Translation}

\author{Kshitij Gupta \\
  Department of Electrical and Electronics Engineering \\
  BITS Pilani, Pilani Campus \\
  Rajasthan, India \\
  \texttt{mailguptakshitij@gmail.com} \\}

\date{}

\begin{document}
\maketitle
\begin{abstract}
Large pre-trained language models have brought remarkable progress in NLP. 
Pre-training and Fine-tuning have given state-of-art performance across tasks in text processing.
Data Augmentation techniques have also helped build state-of-art models on low or zero resource tasks. 
Many works in the past have attempted at learning a single massively-multilingual machine translation model for zero-shot translation. 
Although those translation models are producing correct translations, the main challenge is those models are producing the wrong languages for zero-shot translation. 
This work and its results indicate that prompt conditioned large models do not suffer from off-target language errors i.e. errors arising due to translation to wrong languages. 
We empirically demonstrate the effectiveness of self-supervised pre-training and data augmentation for zero-shot multi-lingual machine translation.
\end{abstract}

\section{Introduction}

Machine Translation is one of the classic problems in Natural Language Processing(NLP).
Several products like Google Translate, Bing Translate provide services to millions of translation requests across a diversity of language pairs.
While the requests for these services come in almost all language pairs imaginable, the quality of translation for low-resource language pairs like German-Arabic is especially low.
This is prompted due to a lack of quality training data for such locale pairs compared to high-resource languages like English-French etc. 

Specifically, for few-shot machine translation, there have been many successful techniques proposed.
\citet{zoph2016transfer} demonstrated that transfer learning from high-resource languages to low-resource languages can be used to achieve remarkably high BLEU scores. 
Building on top of it, \citet{gu2018universal} showed that universal lexical representations achieve better alignment of lexical and syntactic relations between languages.
Similarly, \citep{fadaee2017data} have been successfully used to utilize computer vision leanings in augmentation to low-resource translation.

Success due to techniques like transfer-learning, data augmentation, etc. has also provided great progress in building large multi-lingual neural machine translation models \citep{johnson2017google}.
The objective here is to build a single high-capacity model that is able to generate translations for any language pair and can be trained at the same time.
\citet{zhang2021share} have used conditional specific language routing for achieving impressive performance across low resource language pairs.
Similar to this work, \citet{xia2019generalized} utilized data augmentation strategies and 
\citet{zhang2020improving} used random online back translation to achieve state-of-the-art performance for low-resource machine translation. 
In their work, \citet{arivazhagan2019missing} encourage parameter sharing across language by implementing an auxiliary loss function. 
Similarly, de-noising objective \citep{liao2021improving} and distillation techniques \citep{sun2020knowledge} have also been shown to have boosted zero translation learning. 

Recently, research direction in massively multilingual translation models(MMT) \citep{aharoni2019massively} has also been popularized to build zero-shot translation systems.
\citet{arivazhagan2019massively} provides a survey of challenges associated with MMT models, while also emphasizing the importance of preprocessing and vocabulary in knowledge transfer across language pairs.
Although, \citet{gonzales2020subword} detail the lack of robustness of zero-shot models across training runs, we do not notice it in our training runs and find that augmentation techniques help stabilize the training process.

Neural models like Transformer \citep{vaswani2017attention} have brought significant advances in tasks across NLP. 
Pre-trained language models like BERT \citep{devlin2018bert}, BART \citep{lewis2019bart}, T5 \citep{raffel2019exploring} have achieved state-of-art performance across the NLP spectrum.
Similarly, generative models like GPT-2 \citep{brown2020language} have shown few and zero-shot abilities on many tasks. 
The wide success across tasks has not only been limited to high-resource languages like English, and French but has indeed been shared with low-resource languages like Azerbaijani, Belarussian, Galician, Urdu, etc \citep{lakew-etal-2021-zero}. 
It has also brought significant progress in single large multi-lingual models mBERT \citep{devlin2018bert},  mBART \citep{liu2020multilingual}, mT5 \citep{xue2020mt5} that learn universal representation across languages. 
This is responsible for remarkable zero and few-shot performance across tasks for languages that lack supervised training data. 

In this paper, we will investigate language modeling pre-training and data augmentation strategy for zero-shot translation
Our work provides 2 major and concise pieces of contributions: 
\begin{enumerate}
    \item \textit{Prompt Conditioned} models like mT5 do not suffer from off-target translation and a language tag in the task prompt is sufficient for the model to generate output in the right language.
    \item \textit{SeqMix} style data augmentation technique on top of large pre-trained language models like mT5 is a simple yet competitive approach against a strong baseline on zero-shot translation.
\end{enumerate}

\section{Problem}

We will first look at the challenge of off-target translation. 
For all the zero-shot language pairs, we construct a random test dataset of 1000 examples\footnote{We choose 1000 as a good compromise across languages irrespective of their original test size} from the source language that may or may not be part of the original test dataset. We then run the translation system over those examples and identify all the output that corresponds to the wrong target language. 

Say, a translation model($M$) generates for data instances($x_1...x_n$) translations as ($y'_1...y'_n$) and reference translations as ($y_1...y_n$) for source language($s$) and target language($t$). 
Also, given a language identification oracle as $L$, where $L(x)$ is the predicted language $M$ for data instance$x$.
In this work, we will utilize \citet{salcianu2016cld3}'s Language Identification system to measure language performance. 
We then describe \textit{off-target translation error rate}(OTTER) as: 
\begin{equation}
    OTTER(M) =  \frac{\sum_i L(y'_i) \neq t}{\sum_i L(y_i)=t}     
\end{equation}

In the original paper, \citet{zhang2020improving} used the accuracy of translation language as a metric to compare. 
We argue that any language identification system is noisy and thus accuracy on just translation output doesn't take into account errors of the language identification system. 
OTTER, on the other hand, is a noisy measure that measures language accuracy over both reference and translation output text of the translation system. 
    
The main question that we investigate is improving the quality of zero-resource translation. 
The problem at hand is learning a single model that is able to learn translation across language pairs that are unseen during the training time. 
This is motivated by human language learning experience, that if a person knows German, English, and Arabic and can translate over German $\,\to\,$ English and Arabic $\,\to\,$ English then they should be able to translate with sufficiently good quality between German $\,\to\,$ Arabic without any formal training. 
This is true for us because beneath all the lexical and grammatical differences across languages, we share the grounding of various concepts in the same representation. 
Basically, the representation of the concept `cat' is the same as the word `Katze' in German. 

Traditionally, a pivot language has been prominently used to achieve this task. For example; if a German-to-Arabic translation is required then the original text is passed through the first German-to-English translation engine, and then the output English sentence is passed through the English-to-Arabic translation system.

This simple yet effective strategy has been shown to achieve state-of-art performance across zero-resource settings in many language pairs.
\citet{currey-heafield-2019-zero} use data augmentation with pivot language to generate pseudo-parallel data across zero-shot language pairs and then re-train a system. 
Recently, \citet{dabre2021simultaneous} have even utilized multi-pivot languages and simultaneous translation as a method to improve zero-shot performance.
While \citet{kim2019pivot} combined pivoting with transfer learning and an adapter module, \citet{siddhant2020leveraging} leveraged monolingual data with self-supervision for low resource languages to achieve impressive performance. 

\section{ Experimental Setup}

For the purpose of this study, we will constrain the study and experiments to OPUS-100 by \citet{zhang2020improving}. 
OPUS-100 is an English-centric dataset with over 100 language pairs that have either source/target language as English. 
It also consists of several other non-English-centric pairs that are available for zero-shot translation objectives. 
We should emphasize that while previous and related work on this dataset has been centered around massively-multilingual translation as well as zero-shot translation, the objective of the current work is only on zero-shot translation. 
As part of zero-shot languages, OPUS-100 provides 15 language pairs that are combinations of French, German, Arabic, Russian, Chinese and Dutch. 
For the evaluation of our translation model, we shall use the BLEU score, which is a standard metric of automatic evaluation across machine translation.

We run our experiments using the mT5 implementation available in the Transformers library provided by HuggingFace \citep{wolf2019huggingface} and use pre-trained mT5 models(small, large, and xx-large) from \citet{xue2020mt5}. 
For reproducibility purposes, we use the Adam optimizer and run our experiments on a Google TPU v2-32 instance for 64,000 steps with 256 max length, 512 batch size, 0.0001 learning rate and we use a beam size of 4 at inference time. 

For baseline experiments, we consider 2 strong baselines -- (i) First, for any language pair say XX-YY, we train 2 Transformer models XX-En and En-YY, and run zero shot inference for any new sentence from language XX through XX-En and then through En-YY, note that we do not pre-train these models (ii) Second, we consider the model from \citet{zhang2020improving} which implements random online back translation to recover from off-target translation. 

\section{Methodology}
\subsection{Large Pre-trained Model with Prompt Conditioning}
First, we provide a brief introduction to the T5 architecture. 
T5 or Text-to-Text Transfer Transformer is a recently introduced framework that frames all the NLP tasks as a text-to-text problem. 
While the model architecture is vanilla Transformer architecture \citep{vaswani2017attention}, it has been pretrained on the C4 dataset \citep{raffel2019exploring} on a Masked Language Modeling objective \cite{devlin2018bert}.
Any new task could be provided as a brief prompt to the model along with the input, for example, translation from German to English could be specified as {\tt translate German to English: This is a test input sentence}, while the output is generally not formatted.
This is the default prompt style used by \cite{raffel2019exploring} as well as by this work.  

Recently, \citet{xue2020mt5} introduced the multi-lingual version of this model which is pre-trained on the mC4 dataset. 
They have shown that mT5 exhibits zero-shot capabilities, as learning a task in one language is directly transferable to the same task in a different language without any further training. 
They also highlight that the model suffers from unexpected translation in the output space. 
For example, a model trained on English Part-of-speech and when inference is run on French input outputs an English translation of the French input. 
This issue is similar to what \cite{zhang2020improving} has suggested that the massively multilingual translation models suffer from. 
The lack of language signals to the model results in although correct output but in an incorrect target language. 

\subsection{Data Augmentation}
We run experiments on 2 main techniques in this work:
\begin{itemize}[noitemsep,nolistsep]
    \item \textit{Sentence Concatenation}
    \item \textit{Seq2Mix}
\end{itemize}

\begin{figure*}[t]
    \includegraphics[width=\linewidth]{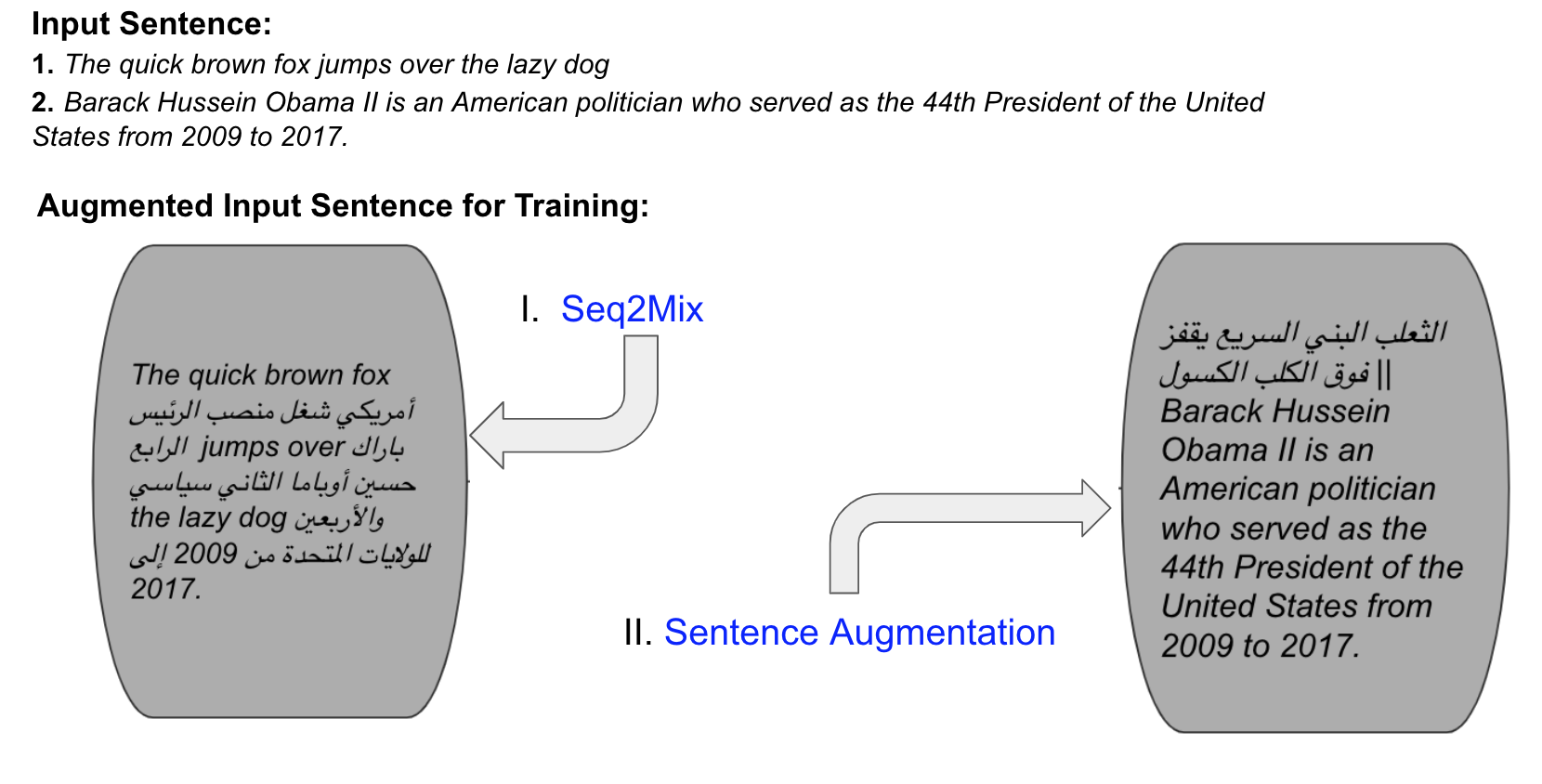}
    \caption{Example of Data Augmentation}
    \label{fig:my_label}
\end{figure*}
In the first set, if our objective is to run translation from German to Arabic, then at each training time, we choose a data point from the German-English dataset and a random data point from the English-Arabic dataset. 
We then concatenate the source sentence and target sentence for both language pairs with a simple \texttt{<sep>} token. 
This results in a training sentence whose input has the first half as a German sentence and the second half as an English sentence. 
Similarly, the output has the first half in English and the Second half as Arabic sentences. 
We modify the input prompt slightly here to identify the languages present in the new input and output, as \texttt{translate German and English to English and Arabic}. 
We hypothesize that the model is intelligent enough to pick up the right words and context from the prompt to produce the right output. 

Secondly, we used Seq2Mix as introduced by \citet{guo-etal-2020-sequence}. 
They propose 2 variants of the Seq2Mix algorithm -- hard and soft Seq2Mix.
For the purpose of this work, we will only focus on the hard version. 
In essence, for 2 input sentences of equal sequence length\footnote{otherwise use padding to ensure the sequences are of equal length}, from German $(Gx_1,...,Gx_n)$ and English $(Ex_1,...,Ex_n)$, we construct a German-English sentence $(GEx_1,...,GEx_n)$, where $GEx_i$ is a token taken randomly from either $(Gx_i, Ex_i)$ with a sample probability from Binomial($\lambda$).\footnote{where $\lambda$ itself is sampled from $\beta$(0.5, 0,5)} 
A similar process is run over to obtain an English-Arabic sentence which serves as output to the German-English input. 

Figure \ref{fig:my_label} depicts both the augmentation techniques for the input sentences in the same. It is noteworthy that we need not merge sentences that are similar to each other, thus we could select any two sentences from our dataset for creating the augmented data.
All of these data augmentation strategies work to create synthetic datasets that are employed along with the original bilingual datasets at training time with equal weighting. 
We hypothesize that while the model learns translation from both synthetic and original datasets, the prompt conditioning along with the mixed vocab is learned better at training time. 

\section{Results \& Analysis}

We find that a mixed data augmentation training regime helps bring down OTTER to an extremely low range as referenced in \ref{tab:otter_metrics}. 
We attribute this to the compositional relationships learned by the large language model on the mixed vocabulary as well as the language tag we use as part of the task prompt.

\begin{table}[!h]
\begin{tabular}{ |c|c|c| } 
 \hline
     & OTTER & BLEU \\ 
 \hline
  Transformer + Pivot & - & 12.98 \\ 
 \hline
  \citet{zhang2020improving} & - & 14.78\\ 
 \hline
  Ours (small mT5) & 27.1\% & 4.9 \\ 
 \hline
  Ours (large mT5) & 24.2\% & 5.1 \\ 
 \hline
  Ours (XXL mT5) & 19.4\% & 7.2 \\ 
 \hline
  XXL mT5 + input concat & 0.9\% & 15.4 \\ 
 \hline
  XXL mT5 + Seq2Mix & \textbf{0.7\%} & \textbf{15.7} \\ 
 \hline
\end{tabular}
\caption{OTTER and BLEU scores for zero-shot language pairs; results are average across all the 15 language pairs in the zero-shot setting}
    \label{tab:otter_metrics}
\end{table}

Similarly, we find that data augmentation techniques like Seq2Mix \cite{guo-etal-2020-sequence}, can substantially improve zero-shot performance when used on top of large pre-trained language models. 
We explain the performance using the following reasoning: 
\begin{itemize}
    \item Mixing vocabulary in the same sentence during training force the internal representation of tokens to align themselves in similar clusters across languages. 
    \item Using XX-English and English-YY translation objective along with data augmentation smoothens the loss landscape to facilitate better representation for zero-shot translation  
\end{itemize}

\section{Conclusion}

In this paper, we utilized mixing augmentation techniques along with large sequence-to-sequence models to generate high-quality zero-shot translation models for language pairs that have no training data available. 
We successfully demonstrate that large pre-trained language models are able to learn the semantic spaces between languages and are already good at zero-shot machine translation. 
However, data augmentation techniques can further boost this performance to achieve impressive results on zero-shot translation. 

\section{Future Work}

We plan on running further experiments with improved data augmentation strategies at pre-training time which we think will benefit downstream zero-shot translation. 

\bibliography{acl2020}
\bibliographystyle{acl_natbib}




\end{document}